\documentclass[conference]{IEEEtran}
\IEEEoverridecommandlockouts
\usepackage{cite}
\usepackage{amsmath,amssymb,amsfonts}
\usepackage{algorithmic}
\usepackage{graphicx}
\usepackage{textcomp}
\usepackage{booktabs} 
\usepackage{multirow}
\usepackage{xcolor}

\def\BibTeX{{\rm B\kern-.05em{\sc i\kern-.025em b}\kern-.08em
    T\kern-.1667em\lower.7ex\hbox{E}\kern-.125emX}}
\begin{document}

\title{HandSplatter: Automated Digital Goniometry from Neural Rendering\\
}

\author{\IEEEauthorblockN{Emmett Chen}
\IEEEauthorblockA{\textit{CAMCA} \\
\textit{Massachusetts General Hospital}\\
Boston, USA \\
ebchen@mgh.harvard.edu}
\and
\IEEEauthorblockN{Neal C. Chen, MD}
\IEEEauthorblockA{\textit{Department of Orthopedic Surgery} \\
\textit{Massachusetts General Hospital}\\
Boston, USA \\
nchen1@mgb.org}
\and
\IEEEauthorblockN{Xiang Li, PhD}
\IEEEauthorblockA{\textit{CAMCA} \\
\textit{Massachusetts General Hospital}\\
Boston, USA \\
xli60@mgh.harvard.edu}
\and
\IEEEauthorblockN{Quanzheng Li, PhD}
\IEEEauthorblockA{\textit{CAMCA} \\
\textit{Massachusetts General Hospital}\\
Boston, USA \\
li.quanzheng@mgh.harvard.edu}
\and
\IEEEauthorblockN{Siyeop Yoon, PhD }
\IEEEauthorblockA{\textit{CAMCA} \\
\textit{Massachusetts General Hospital}\\
Boston, USA \\
syoon5@mgh.harvard.edu}
}

\maketitle
\begin{figure*}[t]
    \centering
    \includegraphics[width=\textwidth, keepaspectratio]{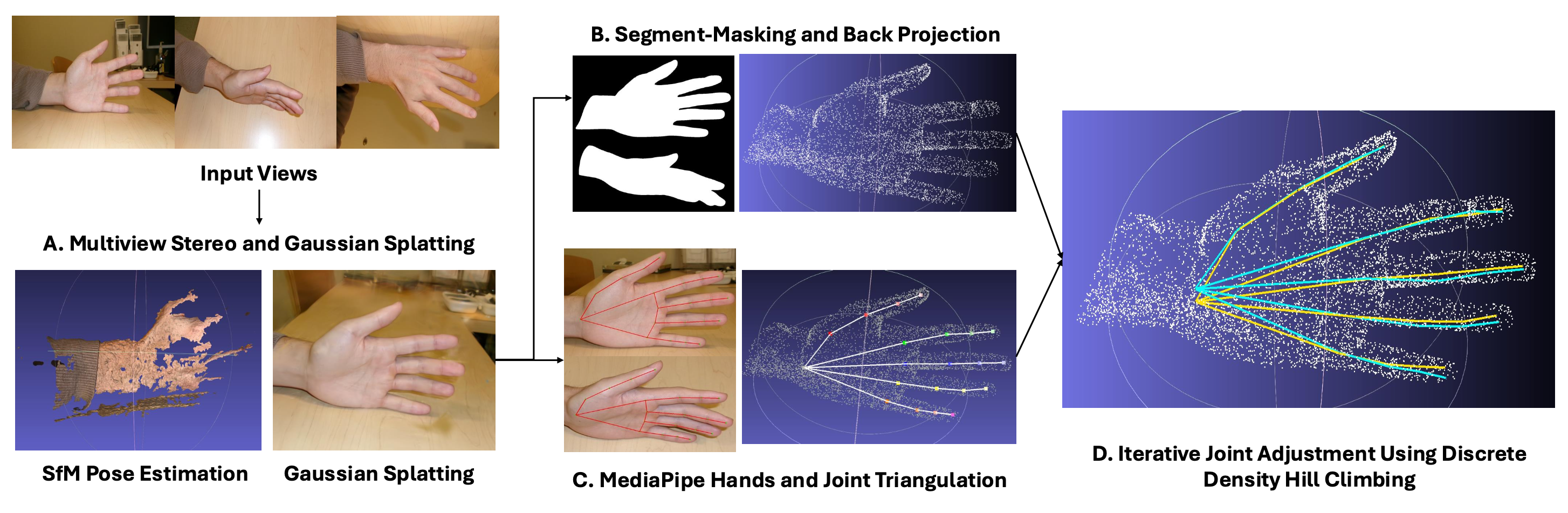}
    \caption{HandSplatter Pipeline Diagram. Shows A. Multiview Stereo and Gaussian Splatting, B. Segment Masking and Back-Projection, C. Mediapipe Hands and Joint Triangulation, D. Iterative Joint Adjustment Using Discrete Density Hill Climbing }
    \label{fig:fullpage}
\end{figure*}

\begin{abstract}
Hand and finger disorders are leading contributors to musculoskeletal disability, creating a clinical need for precise methods to quantify joint motion. Range of motion (ROM) serves as the metric for diagnosis, rehabilitation monitoring, and evaluating surgical outcomes. Currently, the goniometer is the standard tool for assessing finger flexion and extension. However, manual goniometry is labor-intensive and suffers from inconsistent inter-rater reliability due to variations in examiner technique. While digital alternatives exist, current software-based approaches often lack the necessary accuracy for clinical usage.

To address these limitations, we present a novel pipeline for 3-D hand joint location and pose estimation using neural rendering. Unlike previous methods, our approach combines 2-D feature extraction with view synthesis to significantly improve accuracy and clinical viability. Furthermore, we introduce a discrete density hill climbing algorithm that facilitates the meaningful correction of projected landmarks in 3-D space. This system overcomes the inefficiencies of manual measurement and the inaccuracies of existing software, providing a robust tool for objective functional assessment.
\end{abstract}

\vspace{0.1cm}

\section{Introduction}
Hand pose estimation has been well studied in the context of Augmented Reality and Virtual Reality (AR/VR). Recent approaches leverage deep-learning and hand model datasets for accurate monocular pose and depth estimation, such as the transformer-driven HaMeR \cite{pavlakos2024reconstructing} and the relighting-based LiveHand\cite{mundra2023livehandrealtimephotorealisticneural}, which both use the MANO Dataset\cite{MANO:SIGGRAPHASIA:2017} of parameterized hand meshes. In addition, lightweight applications like MediaPipe Hands \cite{zhang2020mediapipehandsondevicerealtime} leverage feature extraction to enable real-time joint and depth estimation on mobile devices \cite{zhang2020mediapipehandsondevicerealtime}. Despite the computational efficiency, the associated loss of anatomical precision renders the clinical effectiveness of these tools uncertain, with evidence suggesting they lack the necessary accuracy for medical use\cite{AMPRIMO2024106508,zhang2020mediapipehandsondevicerealtime}. \\
\indent Beyond monocular pose and depth estimation,  multiview stereo (MVS) methods have been used for full-body and object reconstruction, most notably through neural rendering frameworks such as NeRF \cite{mildenhall2020nerfrepresentingscenesneural} and Gaussian Splatting \cite{KKLD23}, which successfully generate high-fidelity per-scene reconstructions. However, much of the work on neural rendering remains targeted at photorealism and immersive rendering rather than clinical application. \\
\indent In this paper, we present a novel pipeline for 3-D hand joint location and pose estimation using neural rendering. Our approach incorporates Gaussian Splatting\cite{KKLD23} to create a point cloud reconstruction from a set of input views of the hand posed with COLMAP \cite{schoenberger2016mvs, schoenberger2016sfm}. Using Meta's Segment Anything Model (SAM-2) \cite{ravi2024sam2segmentimages} to produce segment masks of the hand, we directionally filter from each camera position, isolating the hand from the background. Next, using the MediaPipe Hands model, we identify joint regions and triangulate them in space. We adjust these joint region locations using a novel discrete density hill climbing algorithm to fit the point cloud. Finally, we measure angles between joints. We evaluated the proposed method against goniometric measurements in n=20 human subjects in resting hand poses.
These results provide preliminary evidence that automated hand pose estimation can approximate traditional clinical assessment while substantially reducing evaluation time.

\section{Methods}

\subsection{Multiview Stereo and Gaussian Splatting}

To obtain a dense and geometrically faithful 3-D representation of the hand suitable for downstream anatomical analysis, we use techniques from multiview stereo and neural rendering. Given a set of calibrated input images ${I_i}_{i=1}^N$, we first apply the COLMAP pipeline \cite{schoenberger2016mvs, schoenberger2016sfm} to perform Structure from Motion (SfM). This yields camera rotational and translational extrinsics ${R_i, t_i}$ with $R_i \in \mathbb{R}^{{3}\times{3}}$ and $t_i \in \mathbb{R}^{{3}\times{1}} $ and intrinsics $K$ for each view, allowing us to model the observed projection $u_i$ of a 3-D point $X \in \mathbb{R}^3$ as

\begin{equation}
    u_i \sim \pi_i(X) = K_i \big(R_i X + t_i\big), \quad u_i \in \mathbb{R}^2 ,
\end{equation}

where $\pi_i$ denotes the perspective projection in view $i$.

Subsequently, we estimate a dense 3-D representation using Gaussian Splatting \cite{KKLD23} with the usage of the Brush 3-D reconstruction engine \cite{brussee2023brush}. Each point is parameterized as a Gaussian ellipsoid
\begin{equation}
    \mathcal{G}_j = \big(\mu_j, \Sigma_j, c_j\big),
\end{equation}
with mean $\mu_j \in \mathbb{R}^3$, covariance $\Sigma_j \in \mathbb{R}^{3\times 3}$, and color $c_j$, producing a continuous radiance field from the posed camera images, $P$.

\subsection{Segment-Masking and Back-Projection}

Gaussian splatting generates a point cloud representation including both the hand and background. We project posed segment masks to isolate the hand. \\
\indent Given an input image $I_k\in \mathbb{R}^{{{H}\times {W}}}$ with camera rotation and translation matrices $(R_k, t_k)$, we segment the image as a mask $M_k$ using a modified SAM-2 segmentation algorithm \cite{ravi2024sam2segmentimages}, where $M_k(u,v) = 1$ for pixels $(u,v)$ belonging to hand and $0$ for background.

Let $K$ be the intrinsic matrix of camera $k$. A pixel $(u,v)$ with depth $d$ back-projects as
\[
X_{k}(u,v,d) = R_k^{\top} \left( K^{-1}\begin{bmatrix} u \\ v \\ 1 \end{bmatrix} d - t_k \right).
\]
We keep points belonging to \[
\mathcal{X} = \left\{ X_k(u,v,d) \;\middle|\; M_k(u,v) = 1 \;\; \forall k \right\}.
\], resulting in a mostly uniform-density point cloud of the hand, with increased density around significant visual features and minimal background.

\subsection{MediaPipe Hands and Joint Triangulation}
After isolating the hand, an initial estimate of joint locations serves as a prior for subsequent refinement. Rather than relying on purely geometric heuristics, we leverage a lightweight, view-consistent 2-D landmark detector to provide coarse joint localization across views.
 Using MediaPipe Hands~\cite{zhang2020mediapipehandsondevicerealtime}, we detect 2-D joints
\[
j_{k,\ell} = f_{\text{MP}}(I_k, \ell), \quad \ell=1,\dots,L,
\]
on each image $I_k$. To loosely locate joints in space, we look for consensus agreement of joint location between different posed hand images. Around each joint we draw a circle
\[
C_{k,\ell} = \{(u,v) \mid \lVert (u,v) - j_{k,\ell} \rVert \leq r\}.
\]

With camera intrinsics $K$ and pose $(R_k,t_k)$, points in $C_{k,\ell}$ back-project to

\begin{equation}
\Pi_{k,\ell}(u,v,d) 
= R_k^{\top}\!\left( K^{-1}
\begin{bmatrix}
u \\ v \\ 1
\end{bmatrix}
d - t_k \right).
\end{equation}
The 3-D joint region is then the set of points $p$ supported by at least 5 images:
\[
\mathcal{J}_\ell = \{\, p \in \mathbb{R}^3 \mid \#\{k : \pi_k(p) \in C_{k,\ell}\} \geq 5 \,\}.
\] of which we only consider the median point.

\subsection{Iterative Joint Adjustment Using Discrete Density Hill Climbing}
Because MediaPipe landmarks are not anatomically constrained, a refinement step is required to align joint estimates with the underlying hand geometry.
Predicted joint regions may fall slightly outside the hand, but may be corrected by iteratively "hill climbing" towards nearby dense regions in the hand point cloud.  

Let $\mathcal{X} = \{x_i \in \mathbb{R}^3\}_{i=1}^N$ denote the normalized hand point cloud,
and let $\hat{j}_\ell^{(t)}$ be the estimate of joint $\ell$ at iteration $t$.

A global hand axis $a \in \mathbb{R}^3$ is obtained as the first principal
component of $\mathcal{X}$. All world coordinates are normalized by the hand’s longitudinal scale, such that the principal axis $a$
has unit length. Distances are expressed in normalized hand-relative units. Joint updates are restricted to directions orthogonal to $a$ to avoid longitudinal displacement of joints along the fingers.
Let $\{d_k\}_{k=1}^K$ be uniformly sampled unit vectors that span the plane
$\{v \in \mathbb{R}^3 : v^\top a = 0\}$ and $\delta = 0.001$ be a fixed step size in normalized world coordinates.

Define a discrete local density at position $x$ as
\[
\rho_r(x) = \sum_{i=1}^N \mathbf{1}\!\left(\|x - x_i\| < r\right),
\]
where $r$ is a fixed radius in normalized world coordinates.

At each iteration, candidate positions
\[
x_k^{(t)} = \hat{j}_\ell^{(t)} + \delta d_k
\]
are evaluated.

Define $k = \arg\max_k \rho_r(x_k^{(t)})$. The update rule is
\[
\hat{j}_\ell^{(t+1)} =
\begin{cases}
\hat{j}_\ell^{(t)} + \delta d_{k}, & \rho_r(x_{k}^{(t)}) \ge m,\\
\hat{j}_\ell^{(t)}, & \text{otherwise}.
\end{cases}
\]

The procedure ends after a fixed number of iterations $it$ or when no candidate direction increases the local density, correcting for MediaPipe's joint misalignment outside of the finger.




\section{Experimentation and Results}

\subsection{Image Collection Protocol}
To obtain sufficient multi-view overlap for 3-D reconstruction, images were collected along a 180° arc around the hand. Subject hands were placed in a resting pose on a flat surface, with ulnar side (small finger) in contact with the table. Camera trajectory began at the palmar view and progressed over the hand to the dorsal view, with 15-30 images captured at regular intervals. This configuration ensured smooth viewpoint transitions and coverage of self-occluding regions.

\subsection{Study Cohort and Exclusion Criteria}

In the final evaluation, we evaluated HandSplatter against manual goniometry on a total of $n=20$ hands. Cases exhibiting
severe self-occlusion, incomplete multi-view coverage, or segmentation
failures were excluded from quantitative analysis, as these conditions
prevented reliable joint localization and angle estimation.

\subsection{Evaluation of Hill Climbing Parameters}
We performed comparative experiments to determine the optimal normalized spherical neighborhood size and number of iterations for the hill climbing algorithm, evaluating a range of $r$ = 0.015–0.03 in normalized world coordinates. Subsequent Bland-Altman analysis revealed that a configuration of r=0.015 with 20 iterations minimized both bias and the limits of agreement versus goniometry (Table \ref{tab:bland_altman}). 


\begin{table}
    \centering
    \caption{Bland Altman Analysis: Goniometer vs. Handsplatter}
    \label{tab:bland_altman}
    \begin{tabular}{l c c c c}
        \toprule
        \textbf{Radius ($r$)} & \textbf{Iterations} & \textbf{Bias} & \textbf{Upper LOA} & \textbf{Lower LOA} \\
        \midrule
        \multirow{3}{*}{0.015} 
            & 10 & 1.00 & 27.53 & -25.52 \\
            & 15 & 1.10 & 27.66 & -25.45 \\
            & 20 & 1.08 & 27.43 & -25.27 \\
        \midrule
        \multirow{3}{*}{0.020} 
            & 10 & 1.41 & 28.86 & -26.02 \\
            & 15 & 1.65 & 29.40 & -26.11 \\
            & 20 & 2.01 & 30.41 & -26.39 \\
        \midrule
        \multirow{3}{*}{0.025} 
            &10 &  1.35 & 29.06 & -26.35 \\
            &15 &  1.64 & 30.15 & -26.86 \\
            &20 &  1.95 & 30.81 & -26.91 \\
        \midrule
        \multirow{3}{*}{0.030} 
            & 10 & 1.34 & 27.95 & -25.26 \\
            & 15 & 1.89 & 28.71 & -24.93 \\
            & 20 & 2.06 & 29.48 & -25.34 \\
        \midrule
        \textbf{Control} & No Adj. & 1.51 & 32.09 & -29.07 \\ 
        \bottomrule
    \end{tabular}
\end{table}

\begin{table}
\caption{Finger by Finger Analysis}
  \label{table:fingerbyfinger}
  \centering
  \begin{tabular}{@{}lc@{}lc@{}lc@{}lc}
    \toprule
    Finger &Bias& &Upper LOA & &Lower LOA&\\
    \midrule
     Thumb     &-0.60 & & 20.82 & & -22.02\\
    Index  & -0.27 & &22.81 && -23.35 \\
    Middle & -0.88 &  & 25.71 & &-27.47 \\
    Ring & 2.48 & & 30.04 & & -25.07\\
    Small  & 5.71 & & 37.27 & & -25.85 \\ 
    \bottomrule
  \end{tabular}
\end{table}

\begin{table}
\caption{Joint by joint analysis}
  \label{table:jointbyjoint}
  \centering
  \begin{tabular}{@{}lc@{}lc@{}lc@{}lc}
    \toprule
    Joint &Bias& &Upper LOA & &Lower LOA&\\
    \midrule
     MCP     &-4.73 & & 16.96 & & -26.43\\
    PIP  & 2.00 & & 24.41 && -24.01 \\
    DIP & 9.01 &  & 36.39 & &-18.37 \\
    Thumb IP & 2.34 & & 24.76 & & -20.09\\
    \bottomrule
  \end{tabular}
\end{table}

\begin{figure}
    \centering
    \includegraphics[width=0.65\linewidth]{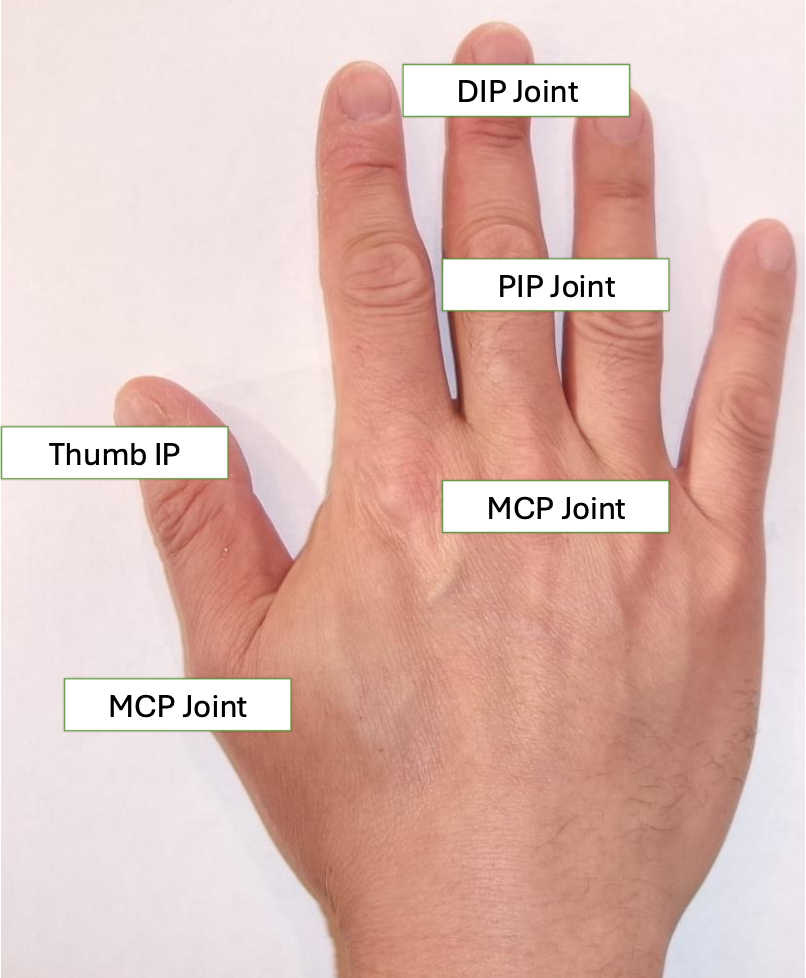}
    \caption{Abbreviated names of finger joints}
    \label{fig:fingerjointnames}
\end{figure}

\subsection{With and Without Hill Climbing Adjustment}

\begin{figure*}[t]
    \centering
    \includegraphics[width=0.38\linewidth]{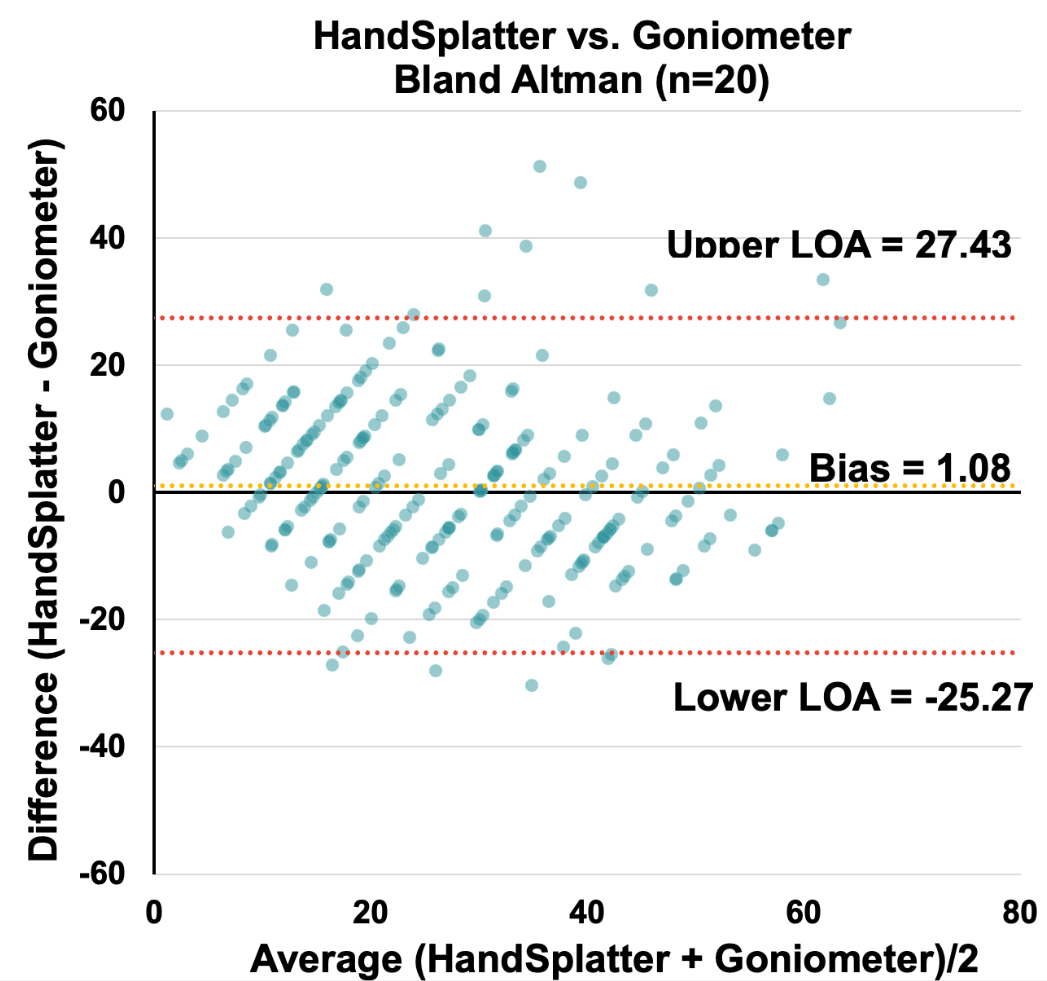}\hfill
    \includegraphics[width=0.38\linewidth]{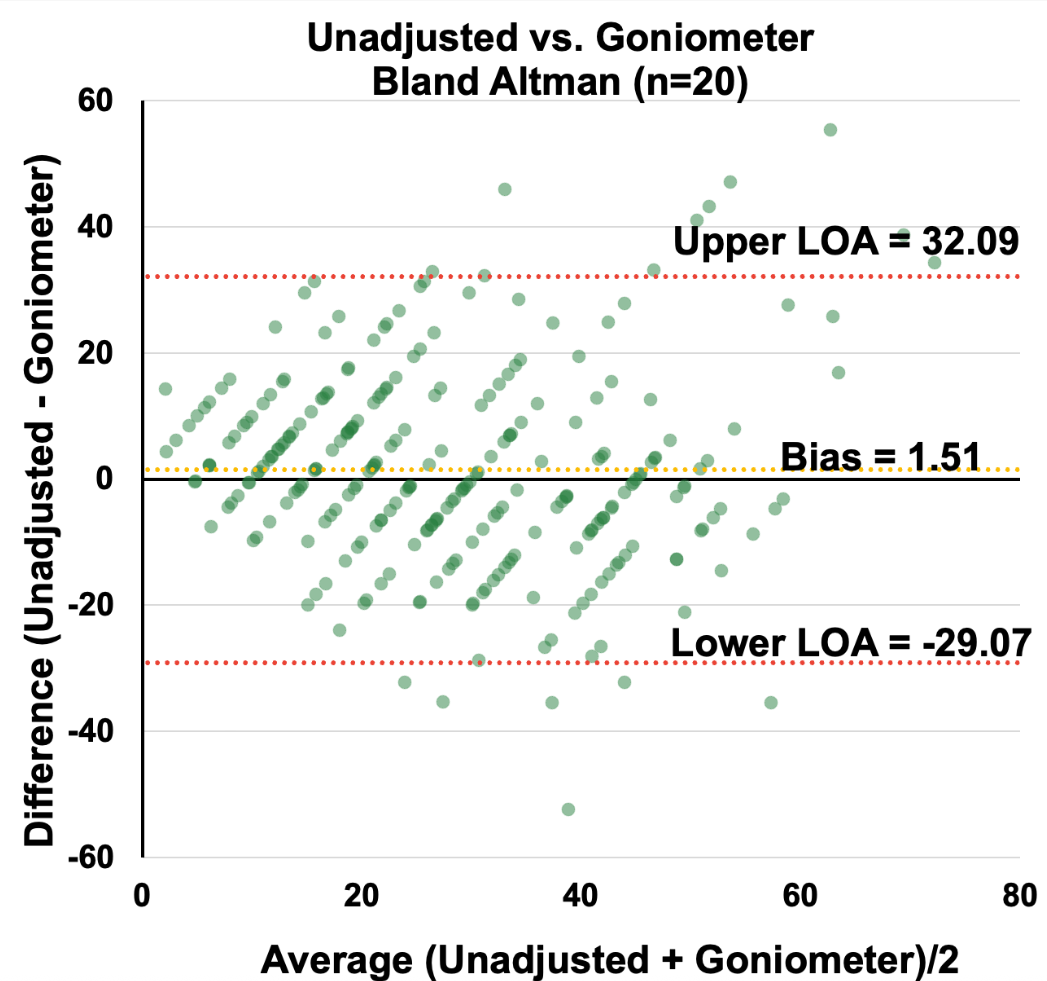}\hfill
    \includegraphics[width=0.2\linewidth]{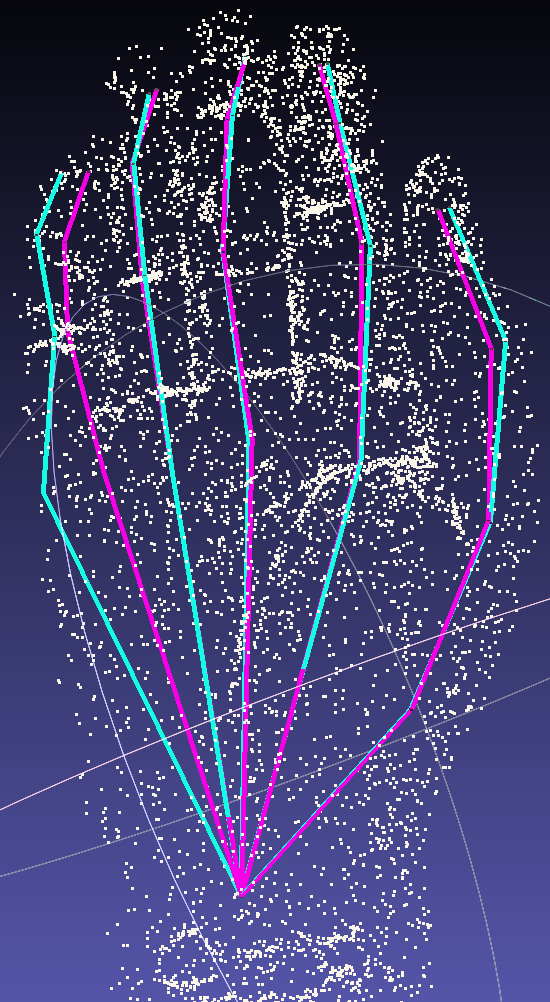}
    \caption{Left: Bland–Altman analysis comparing HandSplatter with iterative adjustment parameters $r=0.015$, $it=20$ against goniometer measurements. Middle: Bland–Altman analysis of HandSplatter without hill climbing adjustment versus goniometry. Right: Qualitative comparison of joint locations with iterative adjustment (magenta) and without adjustment (light blue).}
    \label{fig:band_adjustment_comparison}
\end{figure*}

When using the hill climbing algorithm, range of agreement with the goniometer narrowed by $14\%$ when compared with no adjustment (61.2°) indicating meaningful improvement in correspondence. As illustrated in Figure \ref{fig:band_adjustment_comparison}, this parameter setting also yielded significant qualitative improvements, particularly in the delineation of the joint regions.

\subsection{Per-finger Analysis}

We performed Bland Altman analysis comparing HandSplatter and manual goniometry for each finger, shown in (Table \ref{table:fingerbyfinger}). We found that the thumb, index, middle, and ring fingers had all demonstrated low bias, however the small finger had a bias of +5.71°, indicating moderate systematic overestimation of angles in the small finger. Limits of agreement steadily widened for each finger as one moves from the radial to the ulnar digits (thumb and index to small finger).

\subsection{Per-Joint Analysis}
We also performed Bland Altman analysis comparing HandSplatter and manual goniometry for each joint, shown in (Table  \ref{table:jointbyjoint}). We found that while PIP and thumb IP measurements demonstrated little bias, both MCP and DIP measurements demonstrated notable bias, with MCP measurements biased down -4.73° and DIP measurements biased up +9.01° compared with manual goniometry. Additionally, MCP, PIP, and thumb IP showed substantially narrower ranges of agreement compared to DIP.



\subsection{Discussion, Limitations, and Future Work}

We present a pipeline for 3-D hand joint location and pose estimation that addresses the labor-intensity of manual goniometry as well as the 3-D geometric fidelity limitations of current monocular software-based methods. Neural rendering via Gaussian Splatting enables dense geometric reconstruction of the hand from multi-view imagery captured with handheld mobile devices without reliance on precise camera calibration. 

We also introduce a discrete density hill climbing algorithm that improves robustness by iteratively adjusting joint landmarks towards finger center lines. Both the qualitative and quantitative results indicate meaningful improvement in correspondence (14\% narrower range of agreement) when using hill climbing, supporting the robustness of the proposed approach. More generally, this hill climbing procedure represents a general strategy for correcting imperfect 3-D projections by levering geometric density, enabling improved projection accuracy without the need for precise camera calibration.  

Segmentation quality is a critical element of the proposed pipeline.
Failures most commonly occurred at boundaries with low
chromatic contrast and in regions of severe self-occlusion, particularly
along the ulnar digits during dorsal camera traversal. In these cases,
masking failure led to unreliable hand isolation and joint localization, necessitating exclusion from quantitative analysis. Consistent with this observation, per-finger analysis revealed increasing bias and widening limits of agreement from the radial to ulnar digits, suggesting that reduced visibility is a primary contributor to joint localization error. These findings indicate that image acquisition geometry plays a central role in performance, and improved viewpoint coverage may substantially mitigate these effects. 

Some systematic bias can be attributed to artifact introduced from MediaPipe Hand's anatomical representations. MediaPipe identifies all metacarpal bones as originating from a single point, while in reality, each metacarpal originates from different points in the hand (different carpometacarpal joints). This simplification likely contributes to the observed MCP bias reported in per-joint analysis. Additionally, MediaPipe Hands represents the DIP joint a marker that does not correspond to a well-defined visual anatomical landmark, which may explain the increased bias and limits of agreement observed for DIP measurements. 
Furthermore, manual goniometry itself is subject to variability due to finger motion during measurements and examiner error, and may not necessarily constitute a true anatomical ground truth. These factors should be considered when interpreting systematic differences between automated and manual measurements.   

Despite these limitations, the proposed pipeline is well suited for portable and point-of-care applications. The use of standard handheld imaging devices, absence of specialized hardware, and tolerance to imperfect calibration make the approach practical for deployment in outpatient and inpatient settings, home assessments, and simple longitudinal monitoring

Future work will focus on improving both image acquisition and anatomical modeling. A large arc of views during image collection may increase visibility of ulnar digits, reducing segmentation failure and improving joint localization accuracy. Replacing or augmenting MediaPipe landmarks with more anatomically constrained joint models may improve systematic bias at the MCP and DIP joints. Finally, future studies will evaluate the HandSplatter across broader ranges of hand poses, skin tones, and pathological
conditions, as well as assess intra- and inter-session repeatability to better characterize clinical reliability.

\section{Ethical Statement}
The experimental procedures involving human subjects described in this paper were approved by the Institutional Review Board of Mass General Brigham under protocol number 2025P002242.

\bibliographystyle{IEEEtran}
\bibliography{bibliography}

\vspace{12pt}


\end{document}